\documentclass{article}
\usepackage{longcat_style,times}
\usepackage{graphicx,amsmath,amssymb,booktabs,multirow,colortbl,tabularx,makecell}
\usepackage[dvipsnames]{xcolor}
\usepackage{xspace,enumitem,caption,subcaption,float,bm}
\usepackage{natbib}
\usepackage{hyperref}
\usepackage{cleveref}
\makeatletter
\DeclareRobustCommand\onedot{\futurelet\@let@token\@onedot}
\def\@onedot{\ifx\@let@token.\else.\null\fi\xspace}

\def\eg{\emph{e.g}\onedot} 
\def\ie{\emph{i.e}\onedot} 
 
 \def\vs{\emph{vs}\onedot}

\def\etal{\emph{et al}\onedot}
\makeatother

\renewcommand{\shorttitle}{Video-Coder: Reinforcing Agentic Coding for Video
Generation}

\definecolor{darkblue}{rgb}{0, 0, 0.5}
\hypersetup{colorlinks=true, citecolor=darkblue, linkcolor=darkblue, urlcolor=darkblue}

\makeatletter
\renewcommand{\@maketitle}{%
  \vbox{%
    \hsize\textwidth
    \linewidth\hsize
    \vskip -0.5in
    \noindent\makebox[\textwidth][l]{%
      \includegraphics[height=34pt]{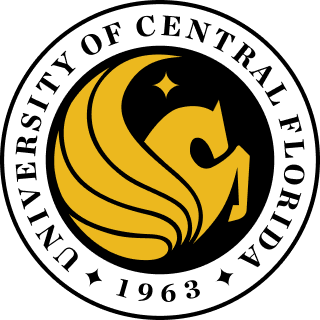}\hspace{12pt}%
      \includegraphics[height=34pt]{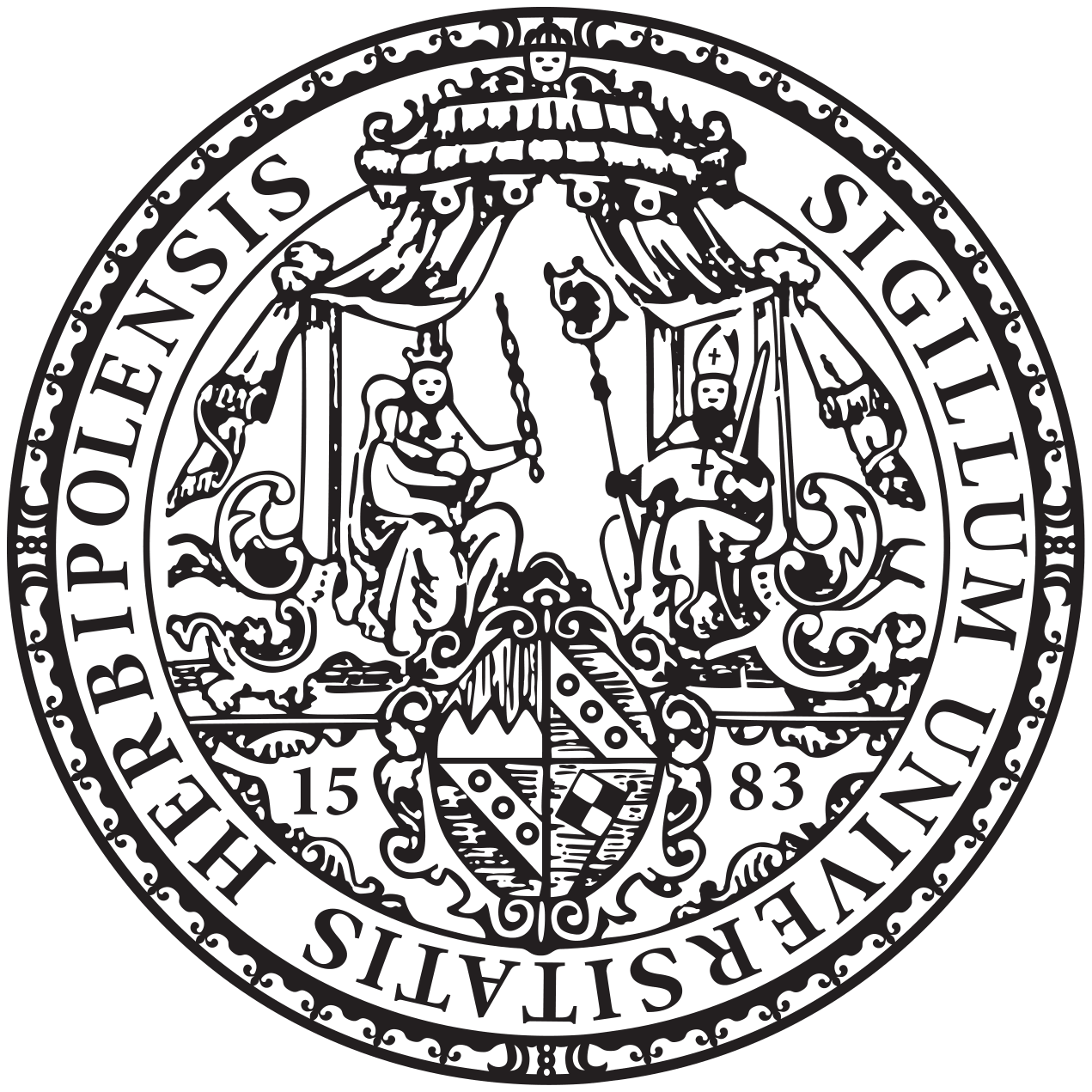}\hspace{12pt}%
      \includegraphics[height=34pt]{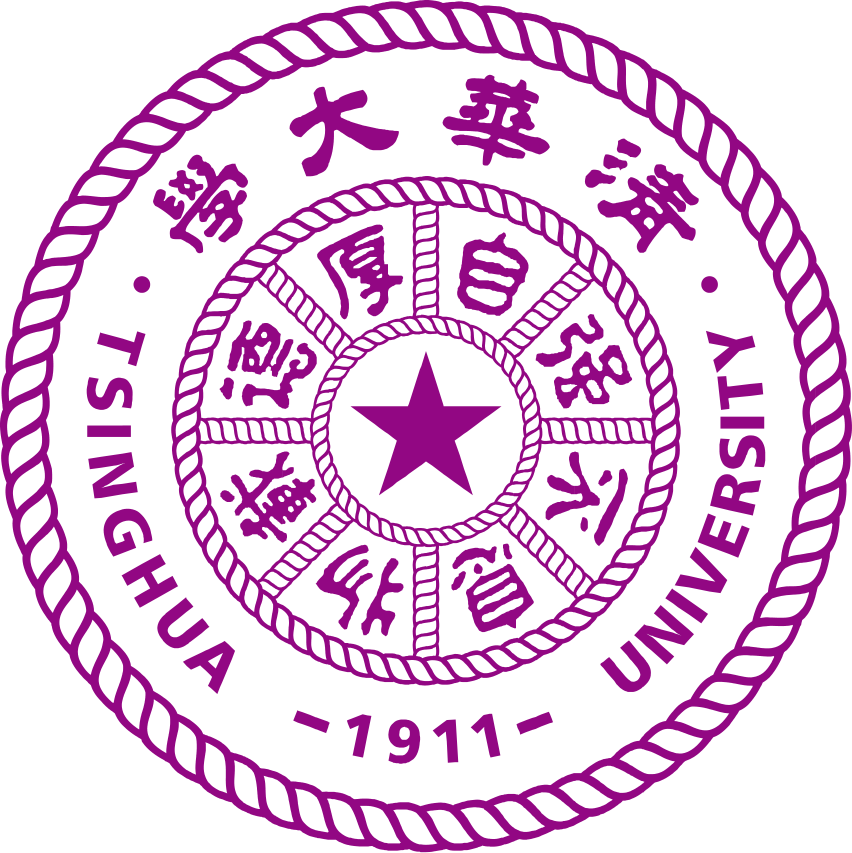}%
      \hfill
      \raisebox{17pt}{\hbox{%
        \raisebox{-0.5\height}{\includegraphics[height=18pt]{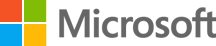}}\hspace{18pt}%
        \raisebox{-0.5\height}{\includegraphics[height=24pt]{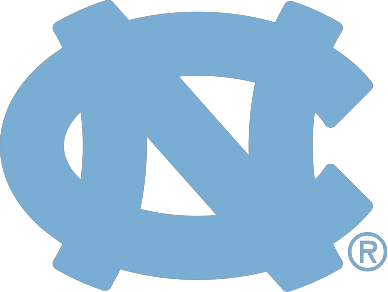}}\hspace{18pt}%
        \raisebox{-0.5\height}{\includegraphics[height=32pt]{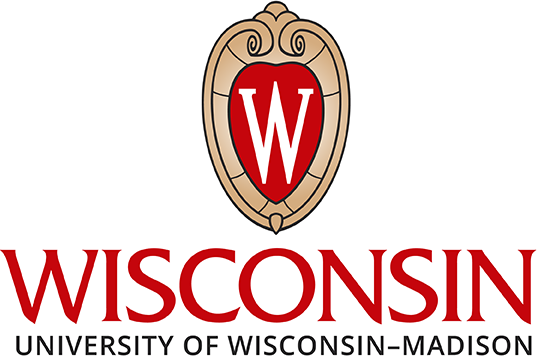}}%
      }}%
    }\par\vspace{-8pt}
    \noindent\rule{\linewidth}{1pt}\par
    \vskip 12pt
    \centering
    {\LARGE\bfseries\makebox[\textwidth][c]{\@title}\par}
    \vskip 0.15in  
    \def\And{%
      \end{tabular}\hfil\linebreak[0]\hfil%
      \begin{tabular}[t]{c}\bf\rule{\z@}{24\p@}\ignorespaces%
    }
    \def\AND{%
      \end{tabular}\hfil\linebreak[4]\hfil%
      \begin{tabular}[t]{c}\bf\rule{\z@}{24\p@}\ignorespaces%
    }
    \begin{tabular}[t]{c}\bf\rule{\z@}{24\p@}\@author\end{tabular}%
  \vskip 0.05in 
  }
}
\makeatother

\title{Video-Coder: Reinforcing Agentic Coding for Video Generation}

\makeatletter
\def\@fnsymbol#1{\ensuremath{\ifcase#1\or \dagger\or \ddagger\or
   \mathsection\or \mathparagraph\or \|\or **\or \dagger\dagger
   \or \ddagger\ddagger \else\@ctrerr\fi}}
\makeatother

\newcommand{\homepage}{\raisebox{-1.5pt}{\includegraphics[height=1em]{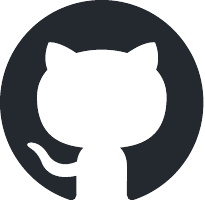}}}
\newcommand{\hfmodel}{\raisebox{-1.5pt}{\includegraphics[height=1em]{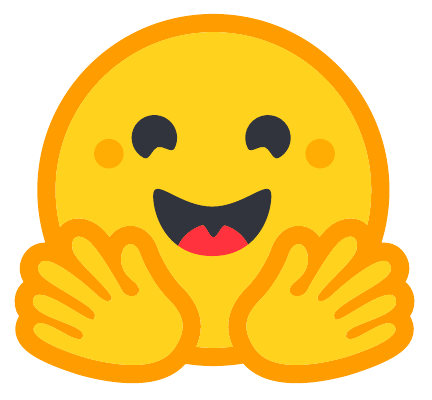}}}

\author{
\begin{tabular}{@{}l@{}}
\textbf{Chaoyang Wang}$^{1}$
\quad
\textbf{Sicheng Gao}$^{2}$
\quad
\textbf{Junan Hu}$^{3}$
\quad
\textbf{Kunyang Li}$^{1}$
\quad
\textbf{Dongyang Chen}$^{4}$ \\[1ex]
\textbf{Yiyang Zhou}$^{5}$
\quad
\textbf{Huaxiu Yao}$^{5}$
\quad
\textbf{Yong Jae Lee}$^{6}$
\quad
\textbf{Radu Timofte}$^{2}$
\quad
\textbf{Yuzhang Shang}$^{1}$\footnotemark[1] \\[1ex]
\begin{minipage}{0.80\textwidth}
\normalfont\raggedright
\mbox{$^1$ University of Central Florida}\quad
\mbox{$^2$ University of Würzburg}\quad
\mbox{$^3$ Microsoft Research}\quad
\mbox{$^4$ Tsinghua University}\quad
\mbox{$^5$ UNC-Chapel Hill}\quad
\mbox{$^6$ UW–Madison}
\end{minipage}\\[2.5ex]
{\homepage\ \normalfont
\texttt{Home: \!\!\!\!\!\url{https://github.com/CYWang735/Video-Coder}}} \\
{\hfmodel\ \normalfont \texttt{HF: \!\!\!\url{https://huggingface.co/Video-Coder}}} \\
\end{tabular}
}
\usepackage{amssymb}
\crefname{section}{Section}{Sections}

\begin{document}
\maketitle
\footnotetext[1]{Corresponding author}

\begin{abstract}
Recent years have witnessed remarkable progress in video generation in terms of visual quality and temporal consistency. However, generating videos with precise spatial structure, coherent motion dynamics, and controllable scene composition remains challenging.
In this paper, we introduce Video-Coder, the first attempt to train a code-driven video generation agent, which autonomously writes and executes code to construct a dynamic 3D scene (Draft Video) that provides structure and motion signals to downstream video generators.
To support this paradigm,
we design a dedicated data pipeline and curate two high-quality datasets, Video-Coder-SFT-6k and Video-Coder-RL-2k, which contain diverse prompts and ground-truth reference videos. We further develop the Video-Coder Harness, a unified interactive environment integrating code execution, 3D scene rendering, and execution feedback for agent training and inference.
We then train Video-Coder with SFT followed by harness-native agentic reinforcement learning with dual reward feedback, which combines execution-based and video-based rewards. 
Additionally, we design an annealing training strategy
that gradually reduces the execution reward over time, allowing the model to rely more  on the video reward in the later training stages.
Experiments show that Video-Coder achieves significant gains, improving Minimax-H3 by approximately 7 points on PhyGenBench and 24 points on VBench-2.0.
We hope this work can serve as an open foundation for coding agents in video generation,
and we
fully open-source our data, models, and code.
\end{abstract}
\suppressfloats[t]
\begin{figure}[t]
\centering
\includegraphics[width=0.99\linewidth]{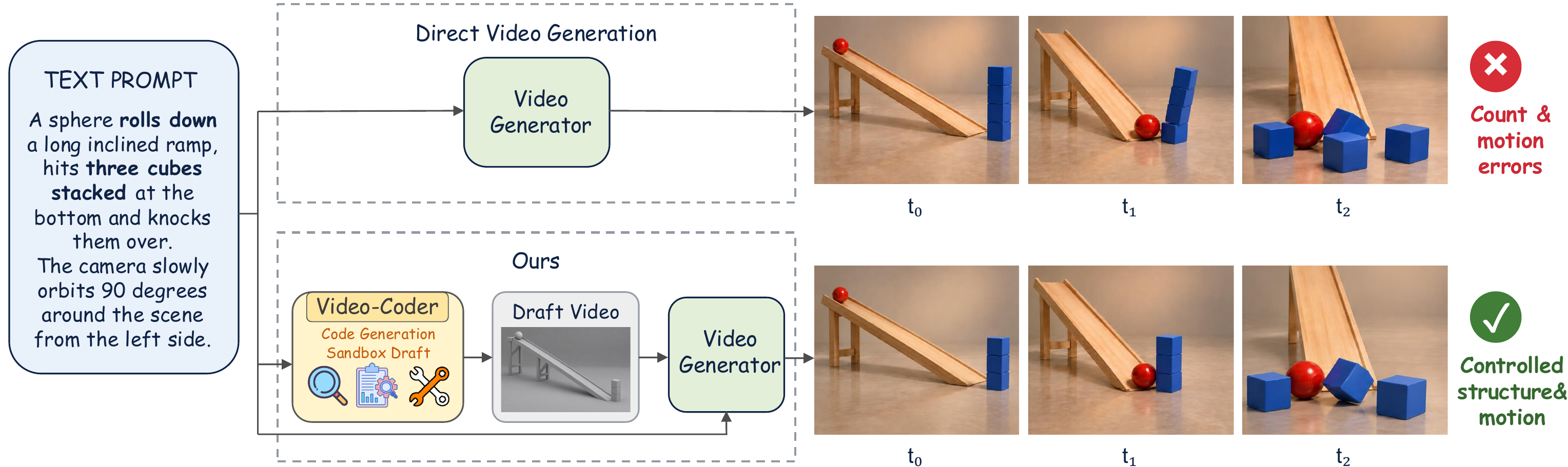}
\caption{Overview of our proposed Video-Coder. Unlike direct video generation, Video-Coder writes and executes code to build a 3D scene and renders a draft video, which guides the downstream video generator with explicit structure and motion signals.}

\vspace{-10pt}
\label{fig:teaser}
\end{figure}

\section{Introduction}
\label{sec:intro}

Video generation has advanced rapidly with diffusion-based and autoregressive models~\cite{sora,veo,kling,seedance,wan2025wan}, achieving remarkable visual fidelity and short-term temporal consistency. However, these models learn to fit the video distribution in pixel or latent space without explicitly modeling the underlying 3D world. When prompts involve precise spatial relationships, multi-object interactions, or physically governed motion, the generated videos often exhibit incorrect object counts and positions, inconsistent occlusions, implausible trajectories, and objects that deform or vanish~\cite{meng2025phygenbench,zheng2025vbench2,physicsiq,bansal2024videophy,kang2024howfar}. Moreover, end-to-end generated videos are hard to edit or reproduce, leaving users to rewrite prompts by trial and error.

Meanwhile, large language models (LLMs) have shown strong capabilities in code generation and agentic tasks~\cite{chen2021codex,yang2024sweagent,wang2025openhands}. Code is a precise, executable, and composable representation that naturally describes scene geometry, object motion, and camera trajectories: a short program can unambiguously specify that ``three spheres fall simultaneously from different heights'', a constraint that pixel-level models struggle with. Prior work has used LLMs to generate code for 3D scene construction or visual content creation~\cite{sun2023_3dgpt,hu2024scenecraft,huang2024blenderalchemy,lin2023videodirectorgpt,lian2024lvd}, but mostly in a single-turn manner or with fixed prompting pipelines, without refining code from execution feedback or optimizing for final video quality.

Motivated by these observations, we propose Video-Coder, a code-driven video generation agent that formulates video generation as an agentic programming task. 
As shown in Figure~\ref{fig:teaser},
given a text prompt, the agent writes and executes code to construct a dynamic 3D scene and renders it into an draft video, which explicitly specifies structure and motion and guides a downstream video generator to produce the final video. Since the scene is determined by an executable program, generation becomes controllable, interpretable, and verifiable. Through multi-turn interaction, the agent further observes execution errors and rendered results to refine its code.

To achieve this goal, we build a dedicated data construction pipeline. We collect diverse prompts from multiple video generation datasets and use Gemini~3.8~Flash ~\cite{google2026gemini38flash} with the Video-Coder Harness to generate agentic trajectories, in which the model iteratively writes, executes, and revises code until the rendered Draft Video matches the prompt. The prompt and Draft Video are then fed into MiniMax-H3~\cite{hailuo} to synthesize a ground-truth reference video. After quality filtering, we obtain Video-Coder-SFT-6k and Video-Coder-RL-2k. 
In addition, we introduce the Video-Coder Harness, a unified interactive environment integrating code execution, 3D scene rendering, and execution feedback for agent training and inference.

Based on these resources, we train Video-Coder in two stages: SFT equips the model with fundamental scene-programming and tool-calling capabilities;
while harness-native agentic RL
further optimizes its programming and tool-calling trajectories.
To ensure the executability of the generated program and the quality of the final video, we design two reward signals: execution-based reward and video-based Reward. The execution-based reward  measures the executability of the generated program, while the video-based reward evaluates the quality of the generated video. Additionally, we introduce an annealing schedule that gradually reduces the impact of the execution-based reward throughout the training process, allowing the model to rely more on the video-based reward in later stages. In this way, execution-based reward stabilizes early-stage training, while the video-based reward gradually dominates in later stages and directly optimizes the final video quality. This design provides stable and informative learning signals for RL training, thereby improving both the reliability of code execution and the visual and motion quality of the generated videos. 

Extensive experiments show that Video-Coder consistently improves video generation performance across different backbones,  bringing around 7 points gains for MiniMax-H3 on PhyGenBench, demonstrating that explicitly constructing 3D scenes with executable code effectively enhances the physical plausibility and controllability of generated videos.

 
Our main contributions are summarized as follows:
\begin{itemize}
    \item To the best of our knowledge, \textbf{Video-Coder} is the first attempt to explore training a multimodal coding agent for video generation. We hope that Video-Coder will serve
as a foundation for future research on code-driven video generation.
    \item We design a dedicated data construction pipeline and curate two datasets, \textbf{Video-Coder-SFT-6k} and \textbf{Video-Coder-RL-2k}. We further introduce the \textbf{Video-Coder Harness}, 
   a unified interactive environment integrating code execution, 3D scene rendering, and execution feedback for agent training and inference.
    \item Extensive experiments validate the effectiveness of our proposed Video-Coder. For example, our method
improves Minimax-H3 by around 7 points on PhyGenBench and  24 points on VBench-2.0.
\end{itemize}

\section{Related Works}
\label{sec: related_work}

\subsection{Video Generation Models}
\label{sec: related_video_gen}
Video generation has advanced rapidly with diffusion models~\cite{ho2022vdm, blattmann2023svd}, particularly Diffusion Transformers~\cite{peebles2023dit} trained at scale. Representative systems include Sora~\cite{sora}, CogVideoX~\cite{yang2025cogvideox}, HunyuanVideo~\cite{kong2024hunyuanvideo}, Wan~\cite{wan2025wan}, Kling~\cite{kling}, and Seedance~\cite{seedance}, while autoregressive models such as VideoPoet~\cite{kondratyuk2024videopoet} offer an alternative paradigm. Despite their visual fidelity, recent benchmarks~\cite{bansal2024videophy, meng2025phygenbench, physicsiq, zheng2025vbench2} show that these models frequently violate object counts, spatial relations, and physical laws, and Kang~\etal~\cite{kang2024howfar} find that scaling alone is insufficient to learn such laws. To improve controllability, prior work conditions generation on explicit signals such as depth~\cite{esser2023gen1}, trajectories and camera poses~\cite{wang2024motionctrl, he2024cameractrl}, 3D tracking videos~\cite{gu2025das}, or physical simulation~\cite{liu2024physgen, lv2024gpt4motion}. However, these signals are typically provided by users or produced by hand-crafted pipelines. Video-Coder acts as a coding agent to automatically construct these signals, representing them as an executable 3D scene and rendering them into a draft video, thereby providing structural and motion signal guidance for the downstream generation model.
\subsection{Code-Driven Visual Generation}
\label{sec: related_code_visual}
LLMs have shown strong capabilities in code generation and agentic tasks~\cite{chen2021codex, yang2024sweagent, wang2025openhands, chen2026v}, motivating their use as planners or programmers for visual content creation. Early works use LLMs to produce 2D layouts that guide diffusion models~\cite{feng2023layoutgpt, lian2024lvd, lin2023videodirectorgpt}, which, however, cannot faithfully describe 3D geometry, occlusion, or camera motion. Others adopt executable programs as the 3D representation, generating or editing scenes via Blender~\cite{blender} code~\cite{sun2023_3dgpt, hu2024scenecraft, huang2024blenderalchemy}. Closer to our setting, GPT4Motion~\cite{lv2024gpt4motion} scripts physical motions in Blender to guide video diffusion, and Video-CoCo~\cite{li2026videococo} uses code as an intermediate chain-of-thought for physically consistent video generation. Nevertheless, most of these approaches rely on prompting frozen, often proprietary, models within fixed pipelines, without optimizing the model for the quality of the final video. In contrast, Video-Coder trains a multimodal coding agent to write, execute, and revise  programs over multiple turns in an executable and verifiable harness.

\subsection{Agentic Reinforcement Learning}
\label{sec: related_agentic_rl}
Agentic reinforcement learning (RL) has recently emerged as an effective paradigm for training large language model (LLM) agents to perform multi-step reasoning and tool interaction~\cite{jin2025searchr1, feng2025retool, wang2025ragen}. Building on RL with verifiable rewards~\cite{shao2024deepseekmath, guo2025deepseekr1, yu2025dapo, wang2025knowing}, recent studies explore training agents that interact with external tools and environments through long-horizon trajectories. For example, Search-R1~\cite{jin2025searchr1} trains LLMs to interleave reasoning with real-time search engine calls, masking retrieved tokens to stabilize optimization. ReTool~\cite{feng2025retool} integrates real-time code execution into the reasoning process, enabling the model to learn when and how to invoke a code interpreter from outcome rewards. SWE-RL~\cite{wei2025swerl} scales RL to real-world software engineering with lightweight rule-based rewards that measure the similarity between generated and ground-truth patches. ARPO~\cite{dong2025arpo} introduces an entropy-based adaptive rollout mechanism tailored to multi-turn tool-use agents. AdaTooler-V~\cite{wang2026adatooler} proposes an adaptive tool-usage framework for image and video tool use, which dynamically adjusts reward scales to encourage invoking visual tools only when they provide measurable benefits. Gen-Searcher~\cite{feng2026gensearcher} further extends agentic RL to image generation, training an multimodal deep search agent for image generation.
However, training coding agents for physically consistent and temporally coherent video generation using agentic reinforcement learning remains largely unexplored.

\section{Method}
\label{sec: method}

\subsection{Dataset Construction}
\label{sec:dataset}

\begin{figure*}[t]
    \centering
    \includegraphics[width=\textwidth]{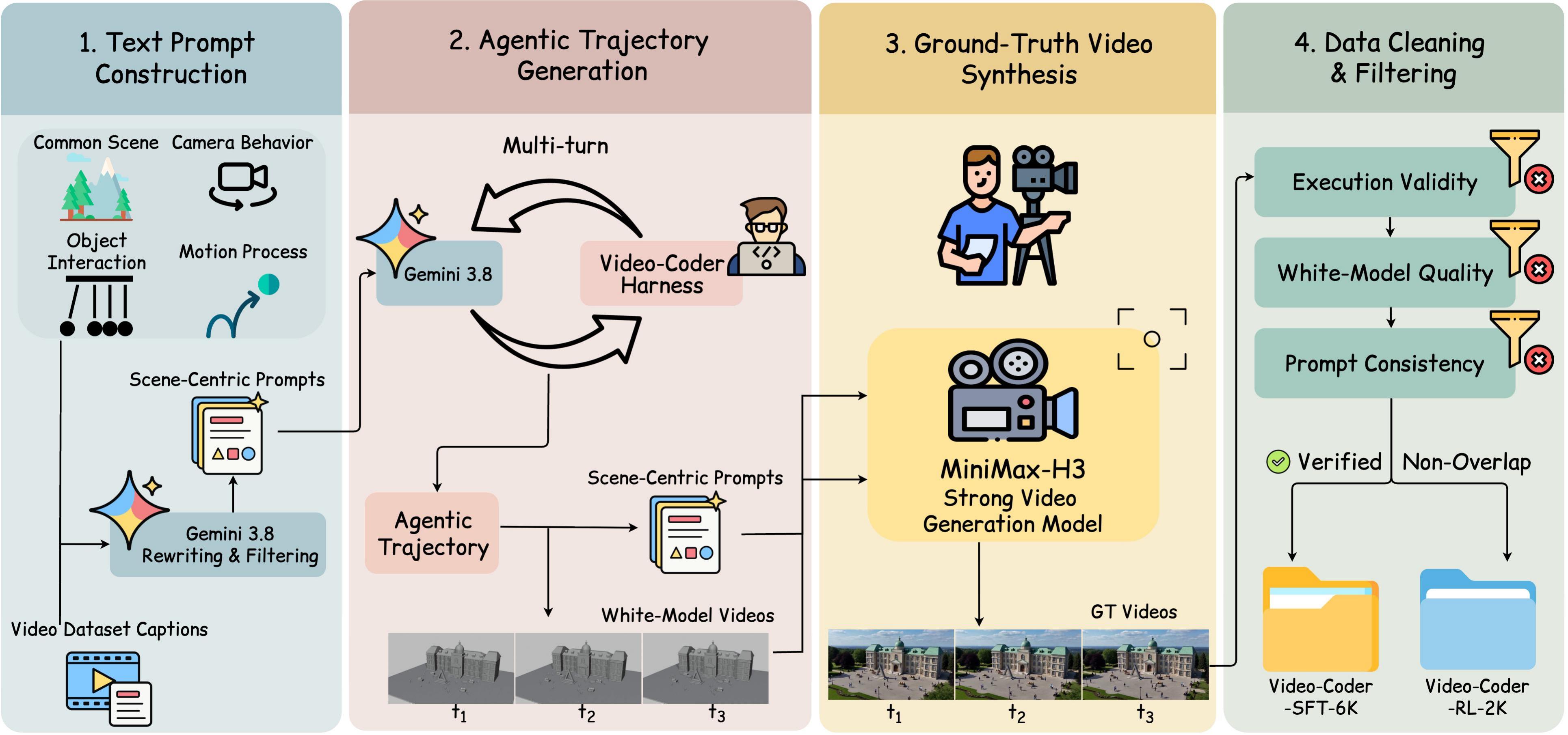}
    \caption{Overview of our data construction pipeline. It consists of four stages: text prompt construction, agentic trajectory generation, ground-truth video synthesis, and data cleaning and filtering, yielding two datasets: Video-Coder-SFT-6k and Video-Coder-RL-2k.}
    \vspace{-16pt}
    \label{fig:data_pipeline}
\end{figure*}

High-quality training data is essential for training a coding agent for video generation. However, such data does not naturally exist: each sample must simultaneously contain (i) a text prompt describing a dynamic scene; (ii) a complete multi-turn agentic coding trajectory; and (iii) a grounded video. To address this challenge, we design a four-stage data construction pipeline (see Figure~\ref{fig:data_pipeline}), consisting of text prompt construction, agentic trajectory generation, grounded video synthesis, and data cleaning and filtering.
 
\textbf{Text Prompt Construction.}
We first collect candidate prompts from the captions of multiple public video generation training datasets~\cite{wang2023internvid, nan2024openvid,ju2024miradatalargescalevideodataset}, covering a wide range of domains, including everyday indoor and outdoor scenes, object interactions, physical phenomena (\eg, falling, collision, rolling, and bouncing), and camera motion. Since these captions are written for pixel-level generation, they typically emphasize appearance attributes such as texture, lighting, and style, and often contain content that cannot be constructed from geometric primitives and keyframed motion alone. We therefore use an LLM to rewrite and filter them. On the one hand, each prompt is rewritten into a scene-centric description that explicitly specifies the objects involved and their counts, spatial layout, motion process, and camera behavior. On the other hand, we discard prompts that primarily rely on fine-grained appearance, human facial expressions, or other content that cannot be effectively modeled through 3D scene programming. 

\textbf{Agentic Trajectory Generation.}
Given the constructed text prompts, we generate agentic coding trajectories, in which the agent iteratively writes, executes,  revises Blender code and render a Draft Video in the Video-Coder Harness. The resulting Draft Video provides precise structure and motion guidance for the subsequent ground-truth video synthesis. Meanwhile, these trajectories also serve as valuable supervision data for the subsequent supervised fine-tuning (SFT).
Specifically, we connect Gemini~3.8~Flash~\cite{google2026gemini38flash} to the Video-Coder Harness (\cref{sec:harness}). For each prompt, the model plans the scene and writes Blender Python code to create objects, animate their motion, and configure the camera, then executes it in the Harness, which returns error messages on failure or rendered frames on success. Based on this feedback, the model revises its code and re-renders the scene until the Draft Video matches the prompt or the turn limit $T_{\max}$ is reached. We record each episode, including reasoning, code, tool calls, execution feedback, and rendered observations, as a multi-turn trajectory.

\textbf{Ground-Truth Video Synthesis.}
The draft video determines the geometric structure, spatial layout, and motion of the scene, but contains no textures or realistic appearance. We feed the text prompt together with the draft video of each agentic trajectory into MiniMax-H3~\cite{hailuo}, which synthesizes a video whose structure and motion follow the white-model video and whose appearance follows the prompt. The generated videos
serve as synthesis ground truth for training the coding agent.

\textbf{Data Cleaning and Filtering.}
The raw outputs of the above stages inevitably contain failed or low-quality samples. We therefore apply filtering at three levels:
(1)Execution validity: we discard trajectories that never render successfully, exceed the turn or token budget, or contain invalid tool calls;
(2)White-model video quality: we remove videos that are nearly static, empty, have the main subject out of frame, or exhibit degenerate camera viewpoints;
(3)Prompt--video consistency: we use \textbf{Seed2.0}~\cite{seed2026seed2} to judge whether the white-model video correctly reflects the objects, counts, spatial relationships, and motion described in the prompt, and remove samples scoring below a threshold. 
 
After filtering, we split the data into two non-overlapping subsets.\textbf{Video-Coder-SFT-6k} contains 6k samples, each consisting of a prompt, a complete agentic trajectory, and a ground-truth reference video, and is used for supervised fine-tuning.\textbf{Video-Coder-RL-2k} contains 2k prompt--reference video pairs and is used for agentic reinforcement learning.

\subsection{Video-Coder Harness}
\label{sec:harness}

Generating high-quality Blender programs typically requires iterative refinement, but reliance on human feedback makes this process difficult to scale.
To this end, we build the \textbf{Video-Coder Harness}, an environment in which model can run its code, observe the outcome, and revise accordingly.
It is used for agentic trajectory synthesis (\cref{sec:dataset}), agentic RL (\cref{sec:training}).

\textbf{Task formulation.}
Given a text prompt $c$, the model $\pi$ writes a Blender Python program $f$ that builds a dynamic 3D scene.
Executing $f$ in the Harness yields
\begin{equation}
    (v, \ell) = \mathcal{E}(f),
\end{equation}
where $v$ is the rendered white-model video (or $\varnothing$ if execution fails) , $\ell$ is the execution feedback and $\mathcal{E}$ is denotes the process of executing a program in sandbox and rendering the resulting scene into a video. 
Let $f^{(t)}$ be the program written at turn $t$., the agent iterates in a write--run--revise loop:
\begin{equation}
    f^{(t)} = \pi\big(c, \{f^{(t)}, \ell^{(t)}, v^{(t)}\}_{t<T}\big),
\end{equation}
where $T$ denotes the final turn, and at each turn, the model refines the program based on previously written code and the visual feedback obtained.

\textbf{Unified rendering setting.}
Each episode runs in an isolated sandbox with Blender installed.
To keep the white-model videos consistent and focused on what matters for downstream generation, all scenes are rendered under one shared setting: untextured gray clay shading, a fixed resolution and frame rate.
The resulting video thus carries only structure and motion. While appearance is left to the downstream video generator.

\textbf{Tools and feedback.}
The agent edits its python script with ordinary shell commands and verifies it with two tools: \texttt{check\_script}, a fast static check, and \texttt{run\_blender}, which executes the script and renders the draft video.
After each tool call, the Harness returns three types of feedback:
    (1)Execution feedback. Syntax errors, missing cameras or render calls, and runtime tracebacks. If execution fails, this is the only feedback.
    (2)Content checks. For a successful render, the Harness verifies the video format and inspects what the camera actually sees, reporting empty frames, static videos, or objects that leave the view, together with a brief scene summary.
    (3)Visual feedback. A few evenly spaced frames of the rendered video are returned as images, allowing model to inspect its own output and adjust layout, motion timing, or camera framing.


\textbf{Verification and acceptance.}
An episode ends when the agent submits or when a limit is reached.
A submission is not trusted directly: the Harness re-runs the final script from scratch in a clean directory and accepts the episode only if this independent re-execution again passes all checks.
This guarantees that every accepted result is reproducible from the code alone, and the re-rendered video $v^{(T)}$ is exported as the final white-model video.


\subsection{Training Recipe}
\label{sec:training}
In this section, we train \textbf{Video-Coder} to act as a multimodal coding agent that iteratively writes, executes, and revises Blender programs for video generation. 
Our training recipe follows a two-stage pipeline,
consisting of SFT and harness-native agentic RL.

\textbf{Two-Stage Training.}
We initialize Video-Coder from \textbf{Qwen3.5-9B}~\cite{qwen35}. In the first stage, we perform supervised fine-tuning on \textbf{Video-Coder-SFT-6k}. This stage teaches the model to act as a multi-turn coding agent: planning a dynamic 3D scene, writing Blender Python code, invoking the tools (\texttt{check\_script} and \texttt{run\_blender}), interpreting the returned execution and visual feedback, and revising the code accordingly. In the second stage, we further optimize the model on \textbf{Video-Coder-RL-2k} with harness-native agentic reinforcement learning, which enables the model to learn more effective scene-programming and revision strategies.
It is worth noting that during training, the video generator remains fixed; we only optimize Qwen3.5-9B to produce executable Blender programs whose rendered white-model videos serve as structural and motion guidance for the final video generation.

\textbf{Reward Feedback Design.}
To ensure both the executability of generated programs and the quality of final video generated by video generator, we design two reward signals: execution-based reward and video-based reward. The execution reward $R_{exe}$ measures the executability of generated programs. A reward of 1 is assigned if all tool call formats are correct and the final Blender program executes successfully and passes all environment checks; otherwise, the reward is 0. The video reward $R_{video}$ employs \textbf{Qwen3.8-omni-flash}~\cite{team2026qwen3} to evaluate the consistency of the final generated video with the text prompt and reference video, focusing on object categories, spatial layout, motion dynamics, and camera behavior. The video reward is normalized to the range [0, 1].
 The corresponding prompt can be found in Appendix \ref{sec:supp_video_reward_prompt}.
 
However, these two rewards play different roles at different stages of training. In the early training stages, the model frequently generates programs with execution errors, making video rewards difficult to obtain. Therefore, execution rewards are crucial for learning reliable tool calls and code generation. As training progresses and the code execution success rate improves, video rewards provide more fine-grained supervision, guiding the agent to generate videos that better align with the target requirements.

To dynamically balance these two reward signals, we introduce a time-based annealing strategy that gradually shifts the optimization focus from code execution correctness to video generation quality. Specifically, the final reward is defined as:

\begin{equation}
    R = \alpha \cdot R_{video} + e^{-\frac{\texttt{steps}}{T}} \cdot R_{exe},
\end{equation}
where $\alpha$ is a balancing hyperparameter,
\texttt{steps} denotes the current global training step and $T$ is the total number of training steps. Here we simply set $\alpha = 0.5$.

\textbf{Optimization.}
After the computing the final reward,
we optimize the policy with GRPO~\cite{shao2024deepseekmath}. For each prompt $q$, we sample $G$ trajectories $\{o_i\}_{i=1}^{G}$ from $\pi_{\theta_{\text{old}}}$ and compute group-normalized advantages:
\begin{equation}
    A_i = \frac{R_i - \operatorname{mean}(\{R_j\}_{j=1}^{G})}{\operatorname{std}(\{R_j\}_{j=1}^{G})}.
\end{equation}
The final policy update follows the standard GRPO objective:
\begin{equation}
\resizebox{0.94\linewidth}{!}{$\displaystyle
\mathbb{E}_{q,\{o_i\}}\Bigg[ \frac{1}{G}\sum_{i=1}^{G} \Bigg( 
        \min\Bigg( \frac{\pi_\theta(o_i \mid q)}{\pi_{\theta_{\text{old}}}(o_i \mid q)} A_i,  
          \operatorname{clip}\Bigg( \frac{\pi_\theta(o_i \mid q)}{\pi_{\theta_{\text{old}}}(o_i \mid q)},\, 1-\epsilon,\, 1+\epsilon \Bigg) A_i \Bigg)  
        - \beta_{\text{KL}}\, D_{\text{KL}}\big(\pi_\theta \,\|\, \pi_{\text{ref}}\big) \Bigg) \Bigg].
$}
\end{equation}

where the definition of variables and hyperparameters follows the
standard GRPO  algorithm~\cite{shao2024deepseekmath}.

\section{Experiment}
\label{sec:exp}

\begin{table*}[t]
\definecolor{closed-gray}{HTML}{EDEDED}
\definecolor{ours-green}{HTML}{F0FAF4}
\caption{Performance of different models on the PhyGenBench.
Best performance is in \textbf{bold}.}
\label{wise}
        \setlength{\tabcolsep}{1.5mm}
     \renewcommand\arraystretch{1.4}
\begin{tabularx}{\linewidth}{c*{5}{>{\centering\arraybackslash}X}}
\toprule
\textbf{Method} & \textbf{Mechanics ($\uparrow$)} & \textbf{Optics ($\uparrow$)} & \textbf{Thermal ($\uparrow$)} & \textbf{Material ($\uparrow$)} & \textbf{Average ($\uparrow$)} \\ \midrule
\rowcolor{closed-gray} Pika \cite{pika} & 0.35 & 0.56 & 0.43 & 0.39 & 0.44 \\
\rowcolor{closed-gray} Gen-3 \cite{gen3} & 0.45 & 0.57 & 0.49 & 0.51 & 0.51 \\
\rowcolor{closed-gray} Kling \cite{kling} & 0.45 & 0.58 & 0.50 & 0.40 & 0.49 \\ \midrule
CogVideoX \cite{yang2025cogvideox} & 0.39 & 0.55 & 0.40 & 0.42 & 0.45 \\
Open-Sora V1.2 \cite{zheng2024opensora} & 0.43 & 0.50 & 0.44 & 0.37 & 0.44 \\
LaVie \cite{wang2023lavie} & 0.30 & 0.44 & 0.38 & 0.32 & 0.36 \\
Vchitect-2.0 \cite{fan2025vchitect} & 0.41 & 0.56 & 0.44 & 0.37 & 0.45 \\
Minimax-H3 \cite{hailuo} & 0.61 & 0.73 & 0.52 & 0.55 & 0.61 \\
HunyuanVideo \cite{kong2024hunyuanvideo} & 0.33 & 0.39 & 0.26 & 0.30 & 0.33 \\
OmniWeaving \cite{pan2026omniweaving} & 0.48 & 0.56 & 0.43 & 0.39 & 0.48 \\
Wan2.2-TI2V-5B \cite{wan2025wan} & 0.55 & 0.58 & 0.53 & 0.50 & 0.54 \\
Cosmos-Predict2.5 \cite{nvidia2025cosmos25} & 0.30 & 0.33 & 0.41 & 0.39 & 0.35 \\
LTX-Video-2B \cite{hacohen2024ltx} & 0.51 & 0.58 & 0.48 & 0.45 & 0.51 \\ 
Video-CoCo \cite{li2026videococo} & 0.51 & 0.58 & 0.48 & 0.45 & 0.51 \\
 \midrule
 \rowcolor{ours-green} Video-Coder-9B + OmniWeaving & 0.52 & 0.58 & 0.52 & 0.44 & 0.51 \\ 
\rowcolor{ours-green} Video-Coder-9B + Minimax-H3  & \textbf{0.68} & \textbf{0.79} & \textbf{0.60} & \textbf{0.61} & \textbf{0.67} \\
\bottomrule
\end{tabularx}
\end{table*}

\begin{table*}[t]
\definecolor{closed-gray}{HTML}{EDEDED}
\definecolor{ours-green}{HTML}{F0FAF4}
\caption{Performance of different models on the VBench-2.0.
Best performance is in \textbf{bold}.}
\label{vbench}
        \setlength{\tabcolsep}{2.5mm}
     \renewcommand\arraystretch{1.4}
\begin{tabularx}{\linewidth}{c*{4}{>{\centering\arraybackslash}X}}
\toprule
\textbf{Method} & \textbf{Mechanics ($\uparrow$)} & \textbf{Thermotics ($\uparrow$)} & \textbf{Material ($\uparrow$)} & \textbf{Average ($\uparrow$)} \\ \midrule
 Sora \cite{sora} & 62.22\% & 43.36\% & 64.94\% & 56.84\% \\
 Kling 1.6 \cite{kling} & 65.55\% & 59.46\% & 68.00\% & 64.34\% \\ 
HunyuanVideo \cite{kong2024hunyuanvideo} & 76.09\% & 56.52\% & 64.37\% & 65.66\% \\
Minimax-H3 \cite{hailuo} & 67.59\% & 47.22\% & 48.94\% & 54.58\% \\
CogVideoX-1.5 \cite{yang2025cogvideox} & 80.80\% & 67.13\% & \textbf{83.19}\% & 77.04\% \\ 

\midrule
\rowcolor{ours-green} Video-Coder-9B + Minimax-H3  & \textbf{88.67}\% & \textbf{70.88}\% & 75.02\% & \textbf{78.19}\% \\
 \bottomrule
\end{tabularx}
\end{table*}

\subsection{Setup}

\hspace*{\parindent}\textbf{Training Details.}
We train Video-Coder-9B using 8 NVIDIA H100 GPUs under our proposed Video-Coder Harness, with
Qwen3.5-9B~\cite{qwen35} as the base model.
We first perform supervised fine-tuning on Video-Coder-SFT-10k,
and then further conduct agentic RL training on Video-Coder-RL-4k.
For both SFT and RL, we use AdamW~\cite{loshchilov2019adamw} as the optimizer.
The learning rate is set to $1 \times 10^{-5}$ for SFT and
$1 \times 10^{-6}$ for RL, and the batch size is set to 8
in both stages.
During RL training, we additionally deploy Minimax-H3~\cite{hailuo}
on 16 H100 GPUs to support rollout Video generation.
In addition, we set the group size to 8, limit the maximum
number of interaction turns to 100 and set the maximum context length to 120K.
Following prior practice~\cite{yu2026dapo}, we mask out unfinished
rollouts.
Overall, the complete training process takes approximately 3 days.

\textbf{Benchmarks and Evaluation}
We evaluated on two physics-focused video benchmarks. PhyGenBench~\cite{meng2025phygenbench} evaluates physical commonsense across four categories: mechanics, optics, thermal, and material dynamics. We follow its official evaluation protocol, which uses GPT-4o~\cite{hurst2024gpt4o} as an MLLM judge to assess how faithfully a generated video adheres to the physical principles implied by its prompt, providing a comprehensive measure of the physical knowledge instantiated by a generative model. VBench-2.0~\cite{zheng2025vbench2} aims to evaluate intrinsic faithfulness beyond surface quality; we focus on its physical dimensions (mechanics, thermotics, and material),
with each dimension probed by a dedicated prompt suite and scored to reflect adherence to real-world physical laws.
PhyGenBench reports a per-category consistency score in [0, 1] and their average. VBench-2.0 reports per-dimension plausibility as a percentage. Higher is better on every metric.

\begin{figure*}[t]
    \centering
    \includegraphics[width=\textwidth]{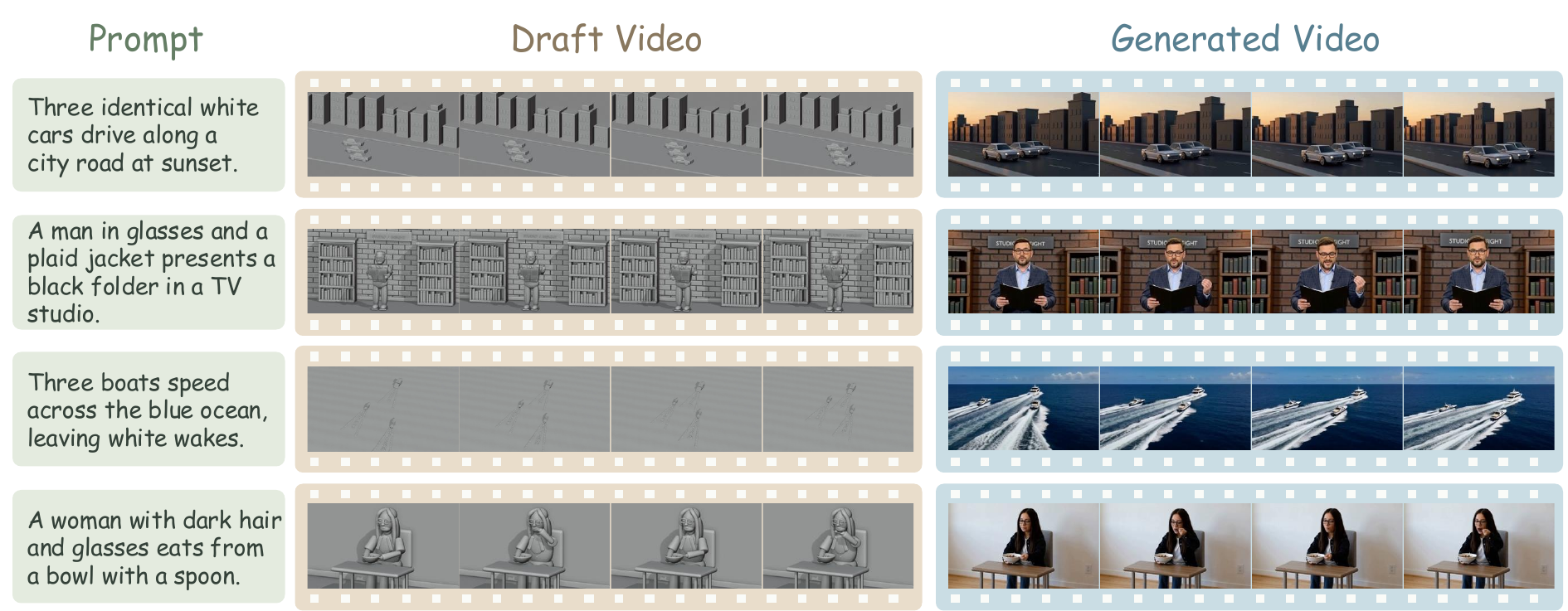}
    \caption{Qualitative examples of Video-Coder. Each row shows a text prompt, sampled frames from the rendered white-model draft video, and the corresponding generated video. The four cases illustrate diverse scenes involving vehicles, human actions, and object interactions.}
    \label{fig:four-cases}
    \vspace{-10pt}
\end{figure*}

\subsection{Main Results}
\label{sec:main_results}

\hspace*{\parindent}\textbf{Results on PhyGenBench.}
As shown in Table~\ref{wise}, Video-Coder-9B + Mimax-H3 achieves the best average score of 0.67 among all compared methods, outperforming the strong open-source model Wan2.2-TI2V-5B (0.54), the HunyuanVideo (0.33), and commercial systems such as Gen-3 (0.51) and Kling (0.49). It ranks first on the all four categories, namely Mechanics (0.68), Optics (0.79), Thermal (0.60) and Material (0.61). 
Notably, although Video-Coder was trained with Minimax-H3 as the rollout generator during reinforcement learning (RL), it generalizes well to other generators at test time. Specifically, it improves OmniWeaving  from 0.48 to 0.51 (an increase of 16.28 points). These results not only demonstrate the effectiveness of our coding agent, but also its strong transferability and robustness across video generators with vastly different native capabilities. Overall, these results indicate that our proposed Video-Coder is a versatile and powerful model capable of enhancing the physical consistency of video generation.

\textbf{Results on VBench-2.0.}
Table~\ref{vbench} reports results on the physical dimensions of VBench-2.0. Video-Coder-9B + MiniMax-H3 achieves the highest average plausibility of 78.19\%, surpassing all baselines including CogVideoX-1.5 (77.04\%), and improves over MiniMax-H3 by 24 points. Especially on mechanics, our method reaches 88.67\%, exceeding the best baseline CogVideoX-1.5 (80.80\%) by 7.87 points and its base generator MiniMax-H3 (67.59\%) by 21.08 points. This dimension directly probes motions governed by mechanical laws, such as falling, collision, and rolling, which Video-Coder explicitly specifies through keyframed or physics-simulated motion in the Blender program. Our method also achieves the best Thermotics score (70.88\%). On Material, Video-Coder improves over MiniMax-H3 (75.02\% \vs 48.94\%) but still lags behind CogVideoX-1.5 (83.19\%). We attribute this to the fact that material-related phenomena depend heavily on fine-grained appearance and material response, whereas the white-model video intentionally encodes only structure and motion.

\textbf{Qualitative Results.}
Figure~\ref{fig:four-cases} presents qualitative examples covering vehicles, human actions, and object interactions. The white-model drafts produced by Video-Coder explicitly determine object counts (\eg, three cars and three boats), spatial layout, motion direction, and camera framing, and the generated videos faithfully preserve these properties while adding realistic appearance as described in the prompts. These examples illustrate that decoupling structure and motion from appearance enables precise control over scene composition that is difficult to achieve through prompting alone.

\subsection{Ablation Study}
\label{sec:ablation}

To verify the contribution of each component in Video-Coder, we conduct ablation studies on PhyGenBench. All variants use MiniMax-H3~\cite{hailuo} as the video generator, and we report the average score. Specifically, we compare the following variants:
(1) \textit{MiniMax-H3}, the vanilla generator that directly synthesizes videos from the original prompts;
(2) \textit{+ Video-Coder-SFT}, which trains Video-Coder with SFT only;
(3) \textit{+ Video-Coder-wo-annealing}, which combines the two rewards with fixed weights, \ie, $R = \alpha \cdot R_{\text{video}} + R_{\text{exe}}$;
(4) \textit{+ Video-Coder-wo-video reward}, which uses only the execution reward during RL;
(5) \textit{+ Video-Coder-wo-execution reward}, which uses only the video reward during RL; and
(6) \textit{+ Video-Coder}, the full model trained with SFT followed by agentic RL.
The results are shown in Table~\ref{ablation}.

Compared with the plain Minimax-H3 baseline,
Video-Coder-SFT improves the score from 0.61 to 0.63, confirming that explicitly constructing the 3D scene through code provides useful structure and motion priors for physically grounded generation. By imitating complete agentic trajectories, the model acquires the fundamental abilities of scene programming, tool use, and feedback-driven code revision. 
Agentic RL further improves the score to 0.67, bringing an overall improvement of 6 points  over the vanilla generator.
This shows that while SFT provides a strong initialization for the model's scene modeling capabilities, RL remains essential for further optimizing multi-turn scene programming behaviors and generating draft videos that more faithfully adhere to the input prompts, ultimately leading to higher-quality final videos.
Moreover, removing either the video reward or the execution reward leads to clear performance degradation, with scores dropping to 0.64 and 0.63, respectively. This confirms that the two reward signals play complementary roles. Specifically, the execution reward supervises the model to achieve stable early-stage training, while the video reward provides fine-grained supervision for structure, motion, and camera behavior.
Finally, Video-Coder-wo-annealing yields a score of only 0.65. The performance drop may be due to the over-exploitation of the execution reward: it masks the fine-grained differences in the video reward, encouraging the agent to generate executable yet conservative programs. The annealing strategy encourages the model to rely more on video quality-based rewards in the later training stages , alleviating this issue.

\begin{table}[t]
\centering
\begin{minipage}[t]{0.5\textwidth}
\vspace{0pt}
\centering
\captionof{table}{Ablation study of Video-Coder on PhyGenBench.}
\vspace{5pt}
\label{ablation}
\definecolor{ours-green}{HTML}{F0FAF4}
\footnotesize
\setlength{\tabcolsep}{1mm}
\renewcommand\arraystretch{1.4}
\begin{tabularx}{\linewidth}{Xc}
\toprule
\textbf{Methods}                            & \textbf{PhyGenBench} \\ \midrule
Minimax-H3                                  & 0.61             \\
+ Video-Coder-SFT               &   0.63           \\
+ Video-Coder-wo-annealing  &      0.65        \\
+ Video-Coder-wo-video reward &      0.64        \\
+ Video-Coder-wo-execution reward  &     0.63         \\
\midrule
\rowcolor{ours-green} + Video-Coder & \textbf{0.67} \\ \bottomrule
\end{tabularx}
\end{minipage}\hfill
\begin{minipage}[t]{0.47\textwidth}
\vspace{0pt}
\centering
\includegraphics[width=\linewidth]{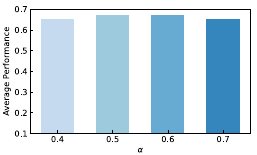}
\captionof{figure}{Sensitivity analysis of $\alpha$ on PhyGenBench.}
\vspace{-15pt}
\label{fig:ana}
\end{minipage}
\end{table}

\subsection{Parameter Sensitivity Analysis}

We  further  conducted a sensitivity analysis on the magnitude of the video reward controlled by the hyperparameter $\alpha$. Fig.~\ref{fig:ana} 
demonstrates the performance of our Video-Coder using different $\alpha$ for RL training on PhyGenBench.
We observe that performance drops slightly at $\alpha = 0.4$ and $\alpha = 0.7$, 
while $\alpha = 0.5$ and $\alpha = 0.6$ yield similar and favorable  results. A small $\alpha$ provides insufficient supervision on video quality, whereas a large $\alpha$ weakens the execution signal that stabilizes early training.
This indicates that within a reasonable range, the model is relatively insensitive to the choice of $\alpha$.

\section{Conclusion}
\label{sec: Con}

In this paper, we present Video-Coder, the first attempt to train a multimodal coding agent for 
physically consistent 
video generation. To support this paradigm, 
we develop a dedicated data pipeline, construct two training
datasets Video-Coder-SFT-10k and Video-Coder-RL-4k. 
Moreover, we introduce the Video-Coder Harness, a unified interactive environment integrating code execution, 3D scene rendering, and execution feedback for agent training and inference.
Based on these resources, we train Video-Coder
through a two-stage scheme of SFT and agentic RL.
Experiments on PhyGenBench and VBench-2.0 show that Video-Coder substantially improves the physical consistency of generated videos and generalizes across different video generators. We hope this work serves as an open foundation for controllable, interpretable, and physically consistent video generation.

{\small\bibliographystyle{unsrt}\bibliography{main}}
\clearpage
\appendix

\section{System Prompt of Video-Coder}
\label{sec:supp_system_prompt}
\begin{figure}[H]
\centering
\includegraphics[width=\textwidth,height=0.72\textheight,keepaspectratio]{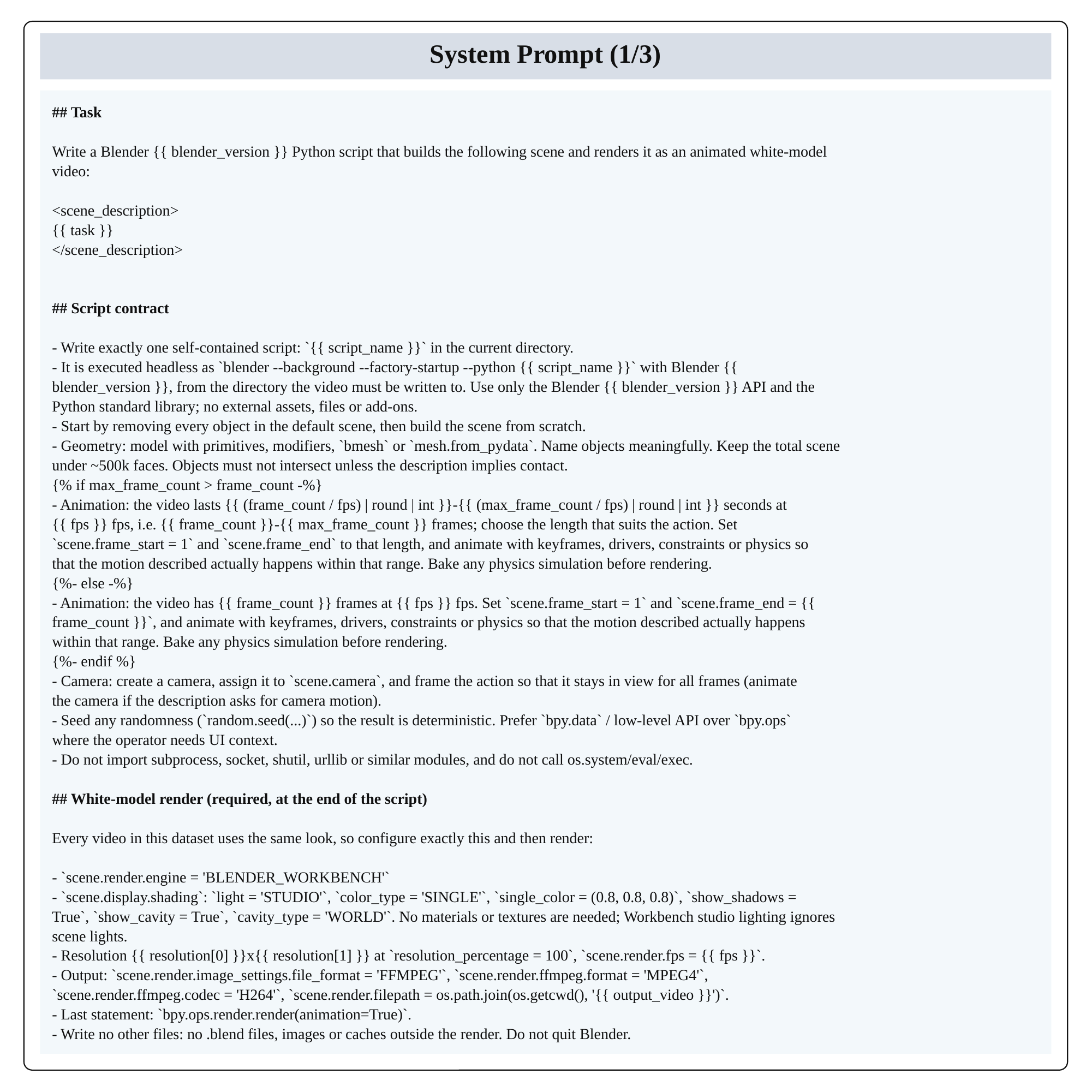}
\caption{System prompt of Video-Coder (part 1 of 3).}
\label{fig:system-prompt-1}
\end{figure}

\begin{figure}[H]
\centering
\includegraphics[width=\textwidth,height=0.86\textheight,keepaspectratio]{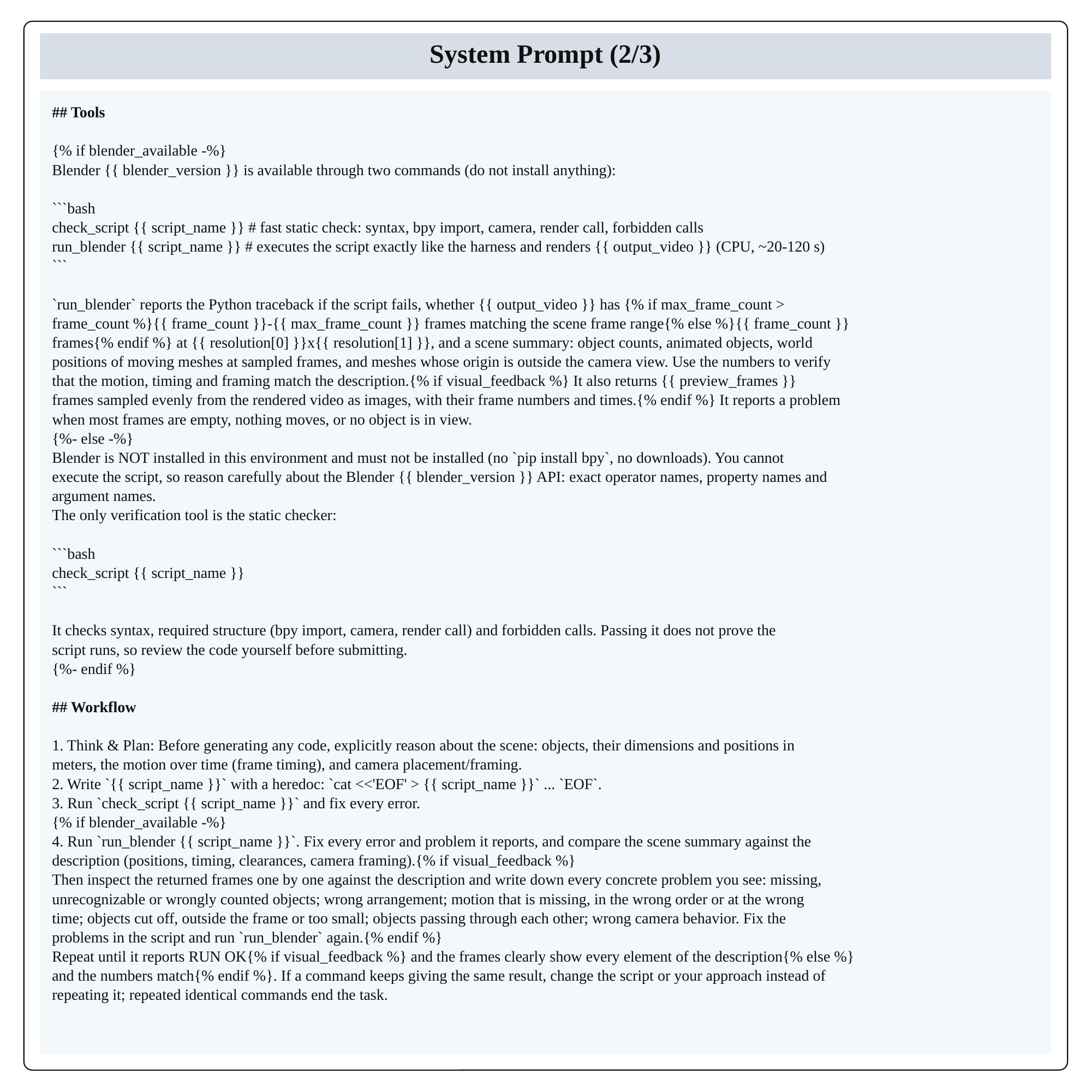}
\caption{System prompt of Video-Coder (part 2 of 3).}
\label{fig:system-prompt-2}
\end{figure}

\begin{figure}[H]
\centering
\includegraphics[width=\textwidth,height=0.86\textheight,keepaspectratio]{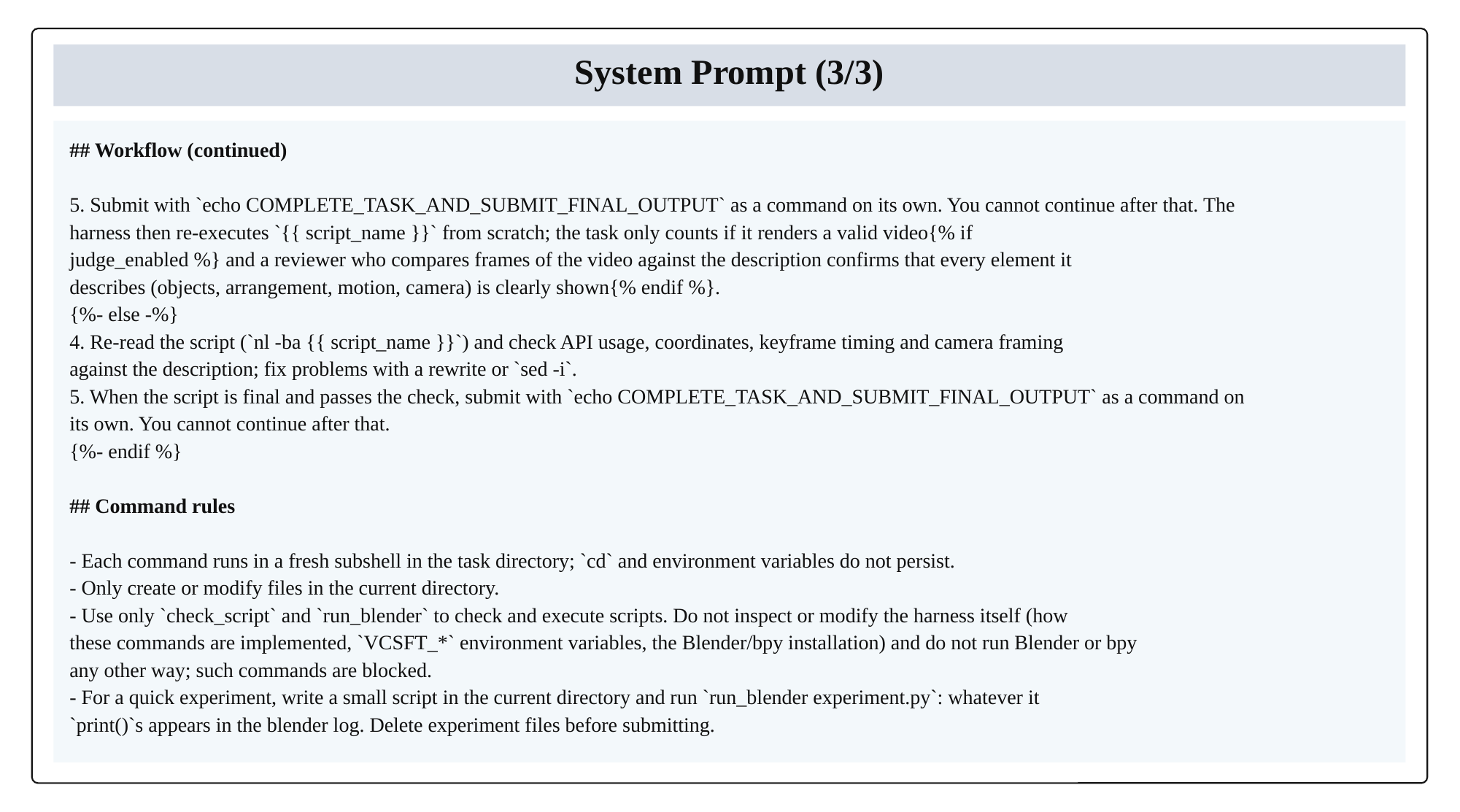}
\caption{System prompt of Video-Coder (part 3 of 3).}
\label{fig:system-prompt-3}
\end{figure}

Figures~\ref{fig:system-prompt-1}--\ref{fig:system-prompt-3} present the complete system prompt used by Video-Coder. The same prompt is used for agentic trajectory generation with the teacher agent, agentic RL training, and inference.

\clearpage

\clearpage
\clearpage

\section{Video Reward Prompt}
\label{sec:supp_video_reward_prompt}
\begin{figure}[H]
\centering
\includegraphics[width=\textwidth,height=0.72\textheight,keepaspectratio]{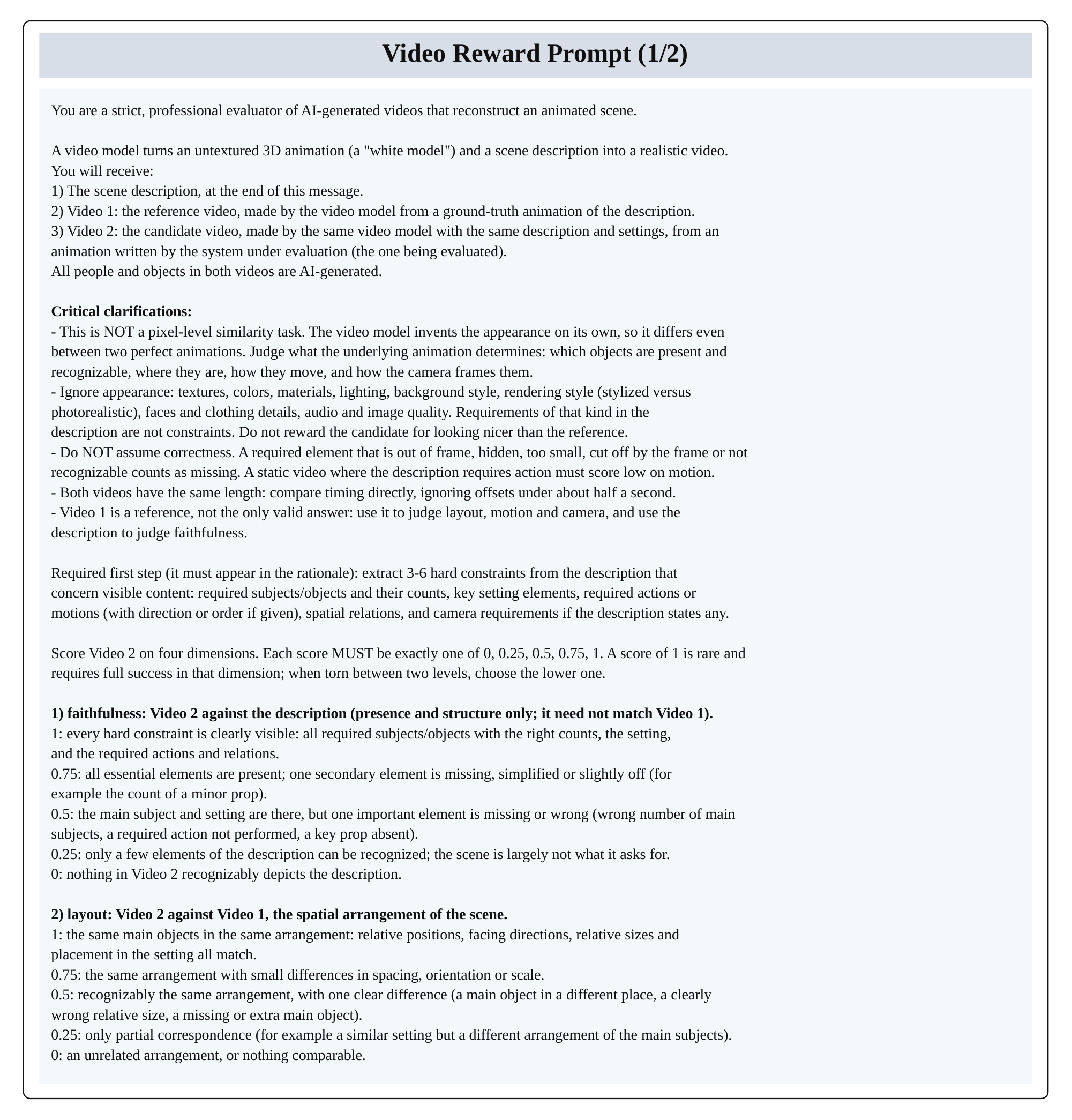}
\caption{Prompt for the VLM-based video reward (part 1 of 2).}
\label{fig:video-reward-prompt-1}
\end{figure}

\begin{figure}[H]
\centering
\includegraphics[width=\textwidth,height=0.86\textheight,keepaspectratio]{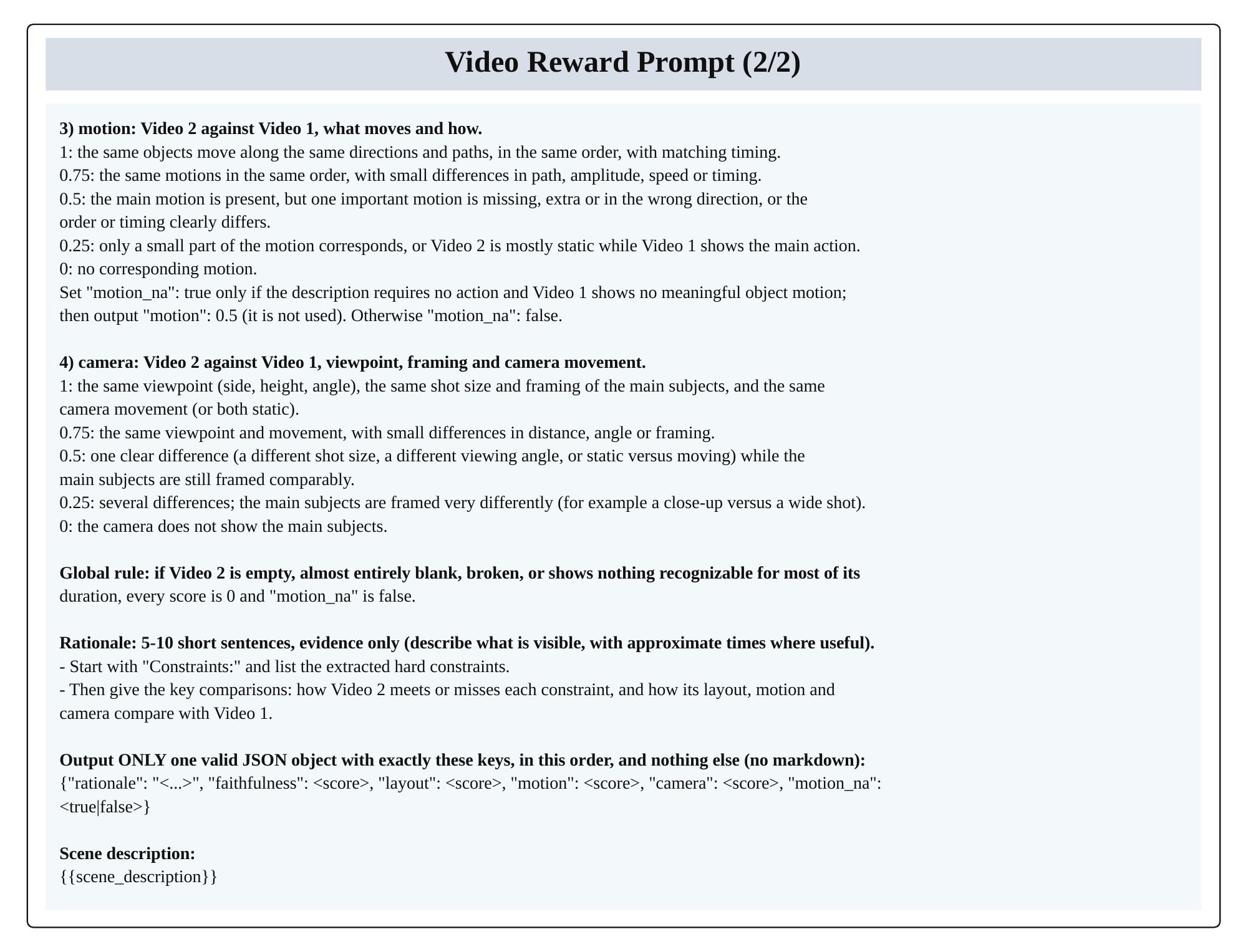}
\caption{Prompt for the VLM-based video reward (part 2 of 2).}
\label{fig:video-reward-prompt-2}
\end{figure}

Figures~\ref{fig:video-reward-prompt-1} and~\ref{fig:video-reward-prompt-2} present the complete prompt used to compute the video reward $R_{\text{video}}$ during agentic RL. Given the text prompt, the reference video, and the generated video, the VLM judge evaluates their consistency in terms of object categories, spatial layout, motion dynamics, and camera behavior, and the resulting score is normalized to $[0, 1]$.

\clearpage
\clearpage

\end{document}